\documentclass[11pt]{article}

\usepackage[preprint]{acl}

\usepackage{times}
\usepackage{latexsym}

\usepackage[T1]{fontenc}

\usepackage[utf8]{inputenc}

\usepackage{microtype}

\usepackage{inconsolata}

\usepackage{graphicx}

\usepackage{amsmath}
\usepackage{amssymb}
\usepackage{booktabs}
\usepackage{multirow}
\usepackage{xspace}
\usepackage{algorithm}
\usepackage{algorithmic}
\usepackage{xcolor}
\usepackage{tcolorbox}
\usepackage{enumitem}
\usepackage{pifont} 

\newcommand{\method}{\textsc{Search-P1}\xspace}

\title{Search-P1: Path-Centric Reward Shaping for Stable and Efficient Agentic RAG Training}

\author{
\textbf{Tianle Xia}$^{*}$, \textbf{Ming Xu}, \textbf{Lingxiang Hu}, \textbf{Yiding Sun}, \textbf{Wenwei Li}, \textbf{Linfang Shang} \\
\textbf{Liqun Liu}$^{\dagger}$, \textbf{Peng Shu}, \textbf{Huan Yu}, \textbf{Jie Jiang} \\
Tencent \\
\texttt{\{tianlexia,flemingxu,lingxianghu,emanuelsun,wenweiwwli,faelynshang\}@tencent.com} \\
\texttt{\{liqunliu,archershu,huanyu,zeus\}@tencent.com} \\
{\small $^{*}$Equal contribution. \quad $^{\dagger}$Corresponding author.}
}

\begin{document}
\maketitle

\begin{abstract}
Retrieval-Augmented Generation (RAG) enhances large language models (LLMs) by incorporating external knowledge, yet traditional single-round retrieval struggles with complex multi-step reasoning.
Agentic RAG addresses this by enabling LLMs to dynamically decide when and what to retrieve, but current RL-based training methods suffer from sparse outcome rewards that discard intermediate signals and low sample efficiency where failed samples contribute nothing.
We propose \method, a framework that introduces \textbf{path-centric reward shaping} for agentic RAG training, comprising two key components: (1) \textbf{Path-Centric Reward}, which evaluates the structural quality of reasoning trajectories through order-agnostic step coverage and soft scoring that extracts learning signals even from failed samples, and (2) \textbf{Dual-Track Path Scoring} with offline-generated reference planners that assesses paths from both self-consistency and reference-alignment perspectives.
Experiments on multiple QA benchmarks demonstrate that \method achieves significant improvements over Search-R1 and other strong baselines, with an average accuracy gain of 7.7 points.
\end{abstract}

\begin{figure}[t]
    \centering
    \includegraphics[width=\columnwidth]{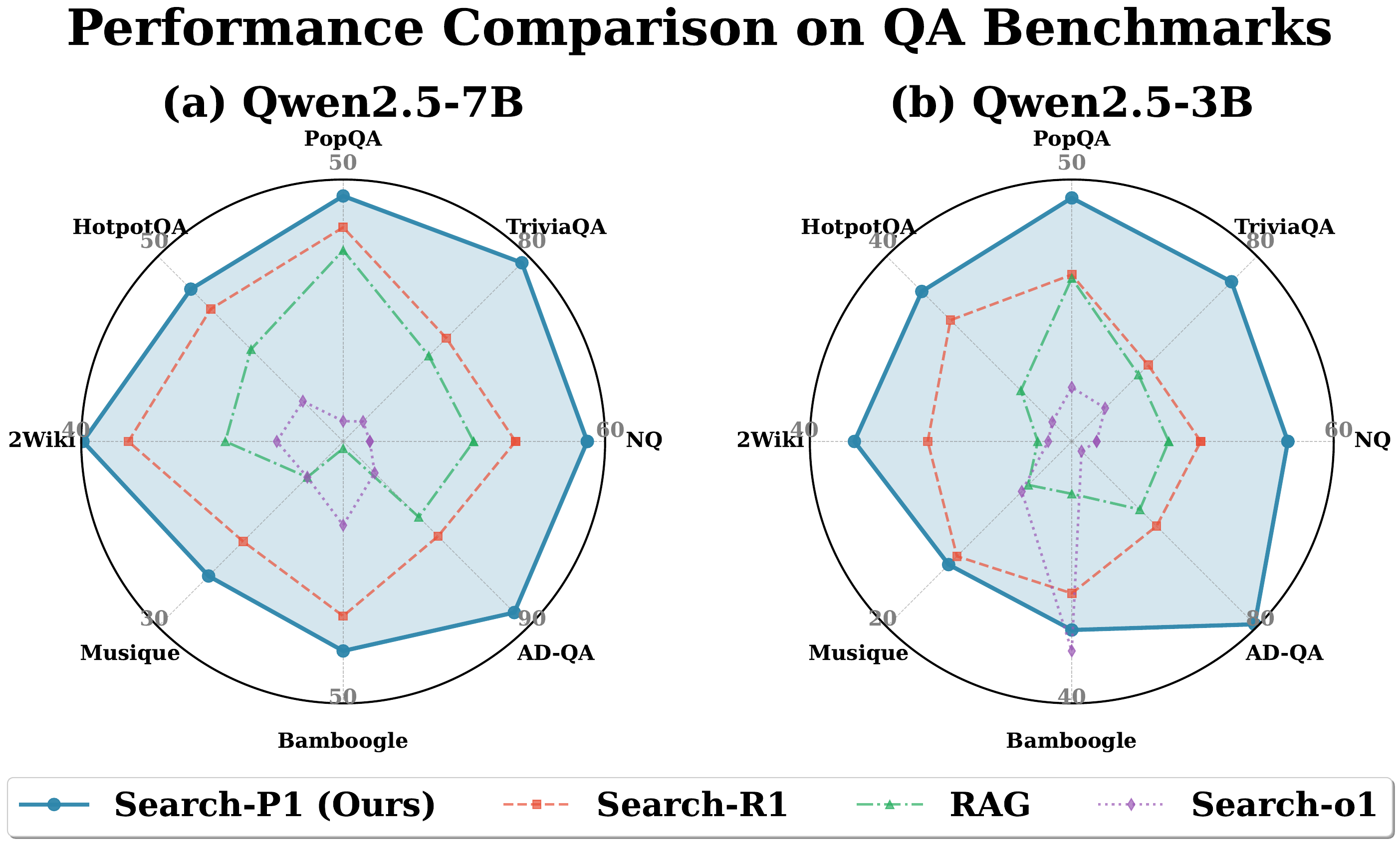}
    \caption{Performance comparison of \method against baselines on QA benchmarks. Our method achieves the highest average accuracy across all datasets on both (a) Qwen2.5-7B and (b) Qwen2.5-3B models.}
    \label{fig:radar}
\end{figure}

\section{Introduction}
\label{sec:introduction}

Large Language Models (LLMs) have demonstrated strong reasoning capabilities~\cite{zhong2023can,xia2025improving,hu2025smarttc}, but their static knowledge often leads to hallucinations on knowledge-intensive queries.
Retrieval-Augmented Generation (RAG)~\cite{10.5555/3495724.3496517} addresses this by incorporating external knowledge, yet single-round retrieval is insufficient for complex multi-step reasoning---a common need in industrial applications such as advertising guidance, where answering a question often requires synthesizing information across multiple knowledge domains.

Agentic RAG extends traditional RAG by enabling LLMs to dynamically invoke search and iteratively refine answers.
Recent methods like Search-R1 apply RL with outcome-based rewards, but this approach has three limitations: (1) \textbf{sparse rewards} that ignore intermediate reasoning quality, (2) \textbf{low sample efficiency} where partially correct trajectories receive zero reward, and (3) \textbf{slow convergence} due to weak training signals when most samples share similar binary rewards.

We propose \method, a framework introducing \textbf{path-centric reward shaping} for agentic RAG training that addresses all three limitations.
Instead of evaluating only final answers, our reward design comprises: (1) \textbf{dual-track path scoring} that provides dense intermediate signals by evaluating reasoning trajectories from both self-consistency and reference-alignment perspectives, directly alleviating reward sparsity; and (2) \textbf{soft outcome scoring} that assigns partial credit to incorrect trajectories, converting zero-reward samples into useful training signals to improve sample efficiency.
Together, the denser reward landscape accelerates convergence by providing more informative gradients throughout training.
Experiments on public QA benchmarks and an internal advertising dataset (AD-QA) show \method outperforms existing methods with an average accuracy gain of 7.7 points, while also transferring effectively to enterprise knowledge base systems.
Our contributions:
\begin{itemize}
    \item We propose dual-track path scoring that evaluates trajectories from self-consistency and reference-alignment perspectives with order-agnostic matching.

    \item We design a path-centric reward shaping framework that extracts learning signals even from failed trajectories via path-level reward.

    \item Extensive experiments on public benchmarks and an industrial dataset demonstrate consistent improvements across models and settings.
\end{itemize}

\section{Related Work}
\label{sec:related_work}

\paragraph{Prompt-Based Agentic RAG.}
Initial efforts leverage prompts to guide LLMs through multi-step retrieval~\cite{singh2025agenticretrievalaugmentedgenerationsurvey,202507.2024}.
These approaches interleave reasoning with retrieval actions~\cite{yao2023react,trivedi-etal-2023-interleaving} or enhance reasoning through sophisticated retrieval strategies~\cite{li2025searcho1agenticsearchenhancedlarge,wang2025chainofretrievalaugmentedgeneration,guan2025deepragthinkingretrievalstep}.
However, prompt-based methods depend heavily on the base model's instruction-following ability.

\paragraph{RL-Based Agentic RAG.}
Recent work applies reinforcement learning to train adaptive search agents~\cite{zhang2025landscapeagenticreinforcementlearning,jin2025searchr1trainingllmsreason}.
Follow-up methods incorporate auxiliary signals to stabilize training~\cite{song2025r1searcherincentivizingsearchcapability,chen2025researchlearningreasonsearch,huang2025ragrladvancingretrievalaugmentedgeneration} or improve search efficiency~\cite{sha2025semreinforcementlearningsearchefficient,song2025r1searcherincentivizingdynamicknowledge,wu2025searchwiselymitigatingsuboptimal}.
Some work explores process rewards for RAG~\cite{10.1145/3726302.3730102,wu2025hiprag,zhang2025process}, but still relies primarily on binary outcome feedback.
Our work proposes path-centric reward shaping offering denser training signals.

\section{Methodology}
\label{sec:method}

\begin{figure*}[t]
    \centering
    \includegraphics[width=\textwidth]{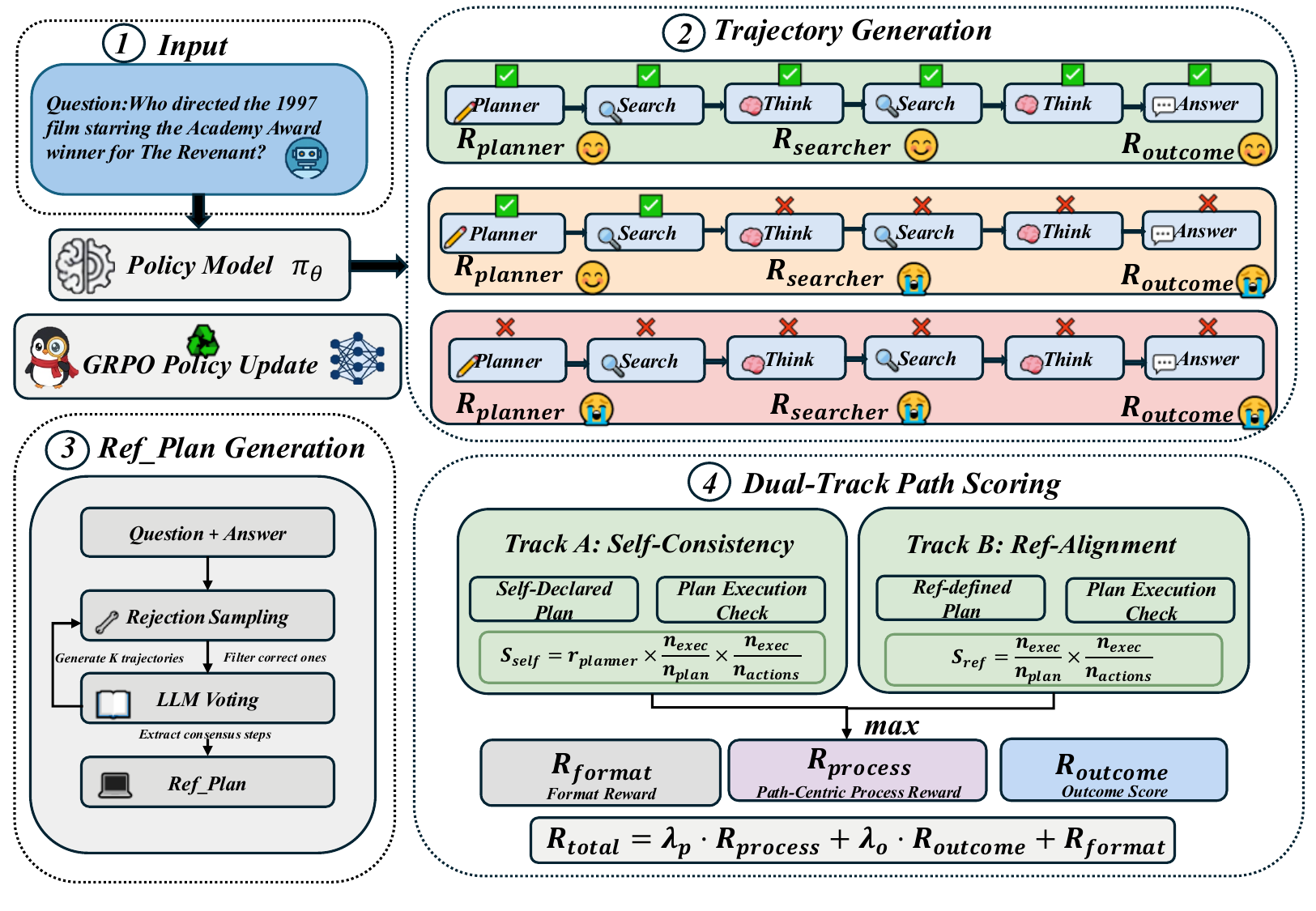}
    \caption{Overview of \method framework. Our approach introduces path-centric reward shaping for agentic RAG training, comprising: (1) Dual-Track Path Scoring that evaluates trajectories from both self-consistency and reference-alignment perspectives, and (2) Soft Outcome Scoring that extracts training signals even from incorrect answers.}
    \label{fig:main}
\end{figure*}

We first formalize the problem setting (\S\ref{ssec:problem}), then describe the path-centric reward framework including dual-track scoring and soft outcome scoring (\S\ref{ssec:path_reward}). Figure~\ref{fig:main} provides an overview.

\subsection{Problem Formulation}
\label{ssec:problem}

We consider an agentic RAG system where a language model $\pi_\theta$ generates a reasoning trajectory $\mathcal{T}$ in response to a question $q$.
In standard agentic RAG frameworks, the trajectory consists of interleaved reasoning and action steps:
\begin{equation}
\mathcal{T} = (r_1, a_1, o_1, \ldots, r_n, a_n, o_n, r_{\text{final}}, \hat{a})
\end{equation}
where $r_i$ denotes reasoning, $a_i$ denotes a search action, $o_i$ is the observation (search results), and $\hat{a}$ is the final answer.

We make the implicit planning in $r_1$ explicit by restructuring the trajectory as:
\begin{equation}
\mathcal{T} = (p, r_1, a_1, o_1, \ldots, r_n, a_n, o_n, r_{\text{final}}, \hat{a})
\end{equation}
where $p$ is an explicit planner that outlines the reasoning strategy. This serves two purposes: (1) providing a self-declared plan against which execution can be evaluated, and (2) making the intended reasoning structure observable for path-centric evaluation.

Standard GRPO assigns binary rewards based on answer correctness:
\begin{equation}
R_{\text{outcome}} = \mathbb{1}[\text{match}(\hat{a}, a^*)]
\end{equation}
where $a^*$ is the ground-truth answer.
This formulation ignores the quality of the reasoning path and suffers from the limitations discussed in \S\ref{sec:introduction}.

\subsection{Path-Centric Reward}
\label{ssec:path_reward}

We propose a path-centric reward that evaluates trajectory quality rather than solely relying on final answer correctness, addressing the three limitations of outcome-based methods. The complete reward function is:
\begin{equation}
R_{\text{total}} = \lambda_p \cdot R_{\text{path}} + \lambda_a \cdot R_{\text{outcome}} + \lambda_f \cdot R_{\text{format}}
\end{equation}
where $R_{\text{path}}$ is the path-centric reward computed via dual-track evaluation, $R_{\text{outcome}}$ is the soft outcome score that extracts signals even from incorrect answers, $R_{\text{format}}$ encourages well-structured outputs, and $\lambda_p$, $\lambda_a$, $\lambda_f$ are balancing coefficients.

\subsubsection{Reference Planner Generation}
\label{sssec:reference}

We generate reference planners offline through rejection sampling and LLM voting.
For each training sample $(q, a^*)$, we generate $K$ candidate trajectories using a high-capability LLM, filter for correct answers, and apply LLM voting to distill an optimized reference planner $P_{\text{ref}}$:
\begin{equation}
P_{\text{ref}} = \text{Vote}(\{T_i\}_{i=1}^{K} | \text{correct}(T_i))
\end{equation}
The voting identifies the minimal set of essential steps across successful trajectories, yielding a reference reasoning path $\mathcal{R}_{\text{ref}} = \{s_1, s_2, \ldots, s_m\}$.

\subsubsection{Dual-Track Path Scoring}
\label{sssec:dual_track}

We evaluate trajectory quality from two complementary perspectives.
Track A (Self-Consistency) assesses whether the model effectively executes its own stated plan:
\begin{equation}
S_{\text{self}} = r_{\text{planner}} \times \frac{n_{\text{exec}}^{\text{self}}}{n_{\text{plan}}} \times \frac{n_{\text{exec}}^{\text{self}}}{n_{\text{actions}}}
\end{equation}
where $r_{\text{planner}}$ rates the plan quality, $n_{\text{exec}}^{\text{self}}$ counts executed steps, $n_{\text{plan}}$ is the total planned steps, and $n_{\text{actions}}$ is the total actions in the trajectory.
Track B (Reference-Alignment) measures coverage of essential steps from the reference planner using order-agnostic matching:
\begin{equation}
S_{\text{ref}} = \frac{n_{\text{covered}}}{|\mathcal{R}_{\text{ref}}|}  \times \frac{n_{\text{covered}}}{n_{\text{actions}}}
\end{equation}
where $n_{\text{covered}}$ counts accomplished reference steps regardless of execution order.
Both tracks incorporate an efficiency ratio $\frac{n_{\text{effective}}}{n_{\text{actions}}}$ to prevent reward hacking through excessive redundant steps and encourage concise reasoning trajectories.
The concrete criteria for determining effective steps and covered steps---including the LLM-based semantic matching procedure---are detailed in Appendix~\ref{sec:appendix_eval_prompt}.
The final path-centric reward $R_{\text{path}} = \max(S_{\text{self}}, S_{\text{ref}})$ takes the maximum rather than a weighted combination, so that when the reference plan is suboptimal or the model discovers a better strategy, the self-consistency track can dominate without being diluted by a low reference score (and vice versa).

\subsubsection{Soft Outcome Scoring}
\label{sssec:soft_outcome}

To improve sample efficiency, we extract learning signals from trajectories with incorrect final answers through soft scoring:
\begin{equation}
R_{\text{outcome}} =
\begin{cases}
1.0 & \text{if correct} \\
\alpha \cdot r_{\text{acc}} + (1-\alpha) \cdot r_{\text{reason}} & \text{otherwise}
\end{cases}
\end{equation}
where $\alpha=0.8$, $r_{\text{acc}}$ indicates partial answer correctness and $r_{\text{reason}}$ evaluates reasoning quality independent of the final answer.
This converts previously zero-reward failed samples into useful training signals based on their path quality.

\section{Experiments}
\label{sec:experiments}

\begin{table*}[t]
\centering
\small
\setlength{\tabcolsep}{5pt}
\renewcommand{\arraystretch}{1.1}
\begin{tabular}{l|ccc|cccc|c|c}
\toprule
\multirow{2}{*}{\textbf{Method}} & \multicolumn{3}{c|}{\textbf{General QA}} & \multicolumn{4}{c|}{\textbf{Multi-Hop QA}} & \multirow{2}{*}{\textbf{Avg.}} & \textbf{Internal} \\
& \textbf{NQ$^\dagger$} & \textbf{TriviaQA} & \textbf{PopQA} & \textbf{HotpotQA$^\dagger$} & \textbf{2Wiki} & \textbf{Musique} & \textbf{Bamboogle} & & \textbf{AD-QA} \\
\midrule
\multicolumn{10}{c}{\textit{Qwen2.5-7B}} \\
\midrule
Direct & 13.4 & 40.8 & 14.0 & 18.3 & 25.0 & 3.1 & 12.0 & 18.1 & 10.3 \\
CoT & 4.8 & 18.5 & 5.4 & 9.2 & 11.1 & 2.2 & 23.2 & 10.6 & 8.7 \\
RAG & 34.9 & 58.5 & 39.2 & 29.9 & 23.5 & 5.8 & 20.8 & 30.4 & 60.4 \\
IRCoT & 22.4 & 47.8 & 30.1 & 13.3 & 14.9 & 7.2 & 22.4 & 23.9 & 52.3 \\
Search-o1 & 15.1 & 44.3 & 13.1 & 18.7 & 17.6 & 5.8 & 29.6 & 20.6 & 48.5 \\
\midrule
Search-R1 & 42.9 & 62.3 & 42.7 & 38.6 & 34.6 & 16.2 & 40.0 & 39.6 & 65.6 \\
HiPRAG & 46.5 & 65.8 & 45.8 & 42.0 & \underline{46.1} & 14.0 & 40.0 & 42.9 & 75.6 \\
\textbf{\method} & \textbf{56.6} & \textbf{78.6} & \textbf{47.5} & \textbf{42.9} & 39.8 & \textbf{21.8} & \textbf{44.0} & \textbf{47.3} & \textbf{86.2} \\
\midrule
\multicolumn{10}{c}{\textit{Qwen2.5-3B}} \\
\midrule
Direct & 10.6 & 28.8 & 10.8 & 14.9 & 24.4 & 2.0 & 2.4 & 13.4 & 7.8 \\
CoT & 2.3 & 3.2 & 0.5 & 2.1 & 2.1 & 0.2 & 0.0 & 1.5 & 5.2 \\
RAG & 34.8 & 54.4 & 38.7 & 25.5 & 22.6 & 4.7 & 8.0 & 27.0 & 54.7 \\
IRCoT & 11.1 & 31.2 & 20.0 & 16.4 & 17.1 & 6.7 & 24.0 & 18.1 & 45.8 \\
Search-o1 & 23.8 & 47.2 & 26.2 & 22.1 & 21.8 & 5.4 & 32.0 & 25.5 & 42.1 \\
\midrule
Search-R1 & 39.7 & 56.5 & 39.1 & 33.1 & 31.0 & 12.4 & 23.2 & 33.6 & 58.3 \\
HiPRAG & 43.0 & 59.8 & 42.0 & 36.0 & \underline{40.5} & 10.8 & 24.0 & 36.6 & 70.2 \\
\textbf{\method} & \textbf{53.0} & \textbf{74.5} & \textbf{47.9} & \textbf{36.2} & 36.6 & \textbf{13.3} & \textbf{28.8} & \textbf{41.5} & \textbf{79.5} \\
\bottomrule
\end{tabular}
\caption{Main results (ACC \%) on seven public QA benchmarks and one internal dataset. Best results are in \textbf{bold}, second best are \underline{underlined}. $^\dagger$ denotes in-domain datasets used for training; others are out-of-domain. AD-QA is a proprietary advertising QA dataset. HiPRAG results are from our reproduction using the same retrieval setup.}
\label{tab:main}
\end{table*}

\begin{table}[t]
\centering
\begin{tabular}{l|c}
\toprule
\textbf{Method} & \textbf{Avg. ACC} \\
\midrule
\method (Full) & 47.3 \\
\quad w/o Reference-Alignment & 42.0 \\
\quad w/o Self-Consistency & 44.2 \\
\midrule
Search-R1 (Baseline) & 39.6 \\
\bottomrule
\end{tabular}
\caption{Ablation study on path-centric reward components (Qwen2.5-7B). Per-dataset results are in Appendix~\ref{sec:appendix_ablation_process}.}
\label{tab:ablation}
\end{table}

\subsection{Experimental Setup}

\paragraph{Datasets.}
Following prior work, we evaluate on seven public QA benchmarks spanning two categories:
(1) \textbf{General QA}: NQ~\cite{kwiatkowski-etal-2019-natural}, TriviaQA~\cite{joshi-etal-2017-triviaqa}, and PopQA~\cite{mallen-etal-2023-trust};
(2) \textbf{Multi-Hop QA}: HotpotQA~\cite{yang-etal-2018-hotpotqa}, 2WikiMultiHopQA~\cite{ho-etal-2020-constructing}, Musique~\cite{trivedi-etal-2022-musique}, and Bamboogle~\cite{press-etal-2023-measuring}.
Additionally, we evaluate on \textbf{AD-QA}, a fully anonymized proprietary advertising QA dataset containing 1,000 multi-hop test instances from an internal business to assess real-world applicability (details in Appendix~\ref{sec:appendix_adqa}).
Following Search-R1, we merge the training sets of NQ and HotpotQA to form a unified training dataset.
Evaluation is conducted on all datasets to assess both in-domain (NQ, HotpotQA) and out-of-domain (TriviaQA, PopQA, 2WikiMultiHopQA, Musique, Bamboogle, AD-QA) generalization.

\paragraph{Models.}
We conduct experiments with Qwen2.5-7B-Instruct and Qwen2.5-3B-Instruct~\cite{qwen2025qwen25technicalreport}, denoted as 7B and 3B for brevity.
For retrieval, we use the 2018 Wikipedia dump as the knowledge source and E5 as the retriever, with top-3 passages returned per search step.

\paragraph{Evaluation Metric.}
We use Accuracy (ACC) as the primary evaluation metric, which checks whether the ground-truth answer is contained in the model's generated response.

\paragraph{Baselines.}
We compare against the following methods:
(1) \textbf{Direct Inference}: Generation without retrieval, including direct prompting and Chain-of-Thought (CoT);
(2) \textbf{Standard RAG}: Single-round retrieval before generation;
(3) \textbf{Prompt-Based Agentic RAG}: IRCoT and Search-o1 that use prompting for multi-step retrieval;
(4) \textbf{RL-Based Agentic RAG}: Search-R1 and HiPRAG that use reinforcement learning  for training.
All RL-based methods share identical training and retrieval configurations (detailed in Appendix~\ref{sec:appendix_impl}); the only difference is the reward function.

\subsection{Main Results}

As shown in Table~\ref{tab:main}, \method achieves the highest average accuracy across both model sizes, outperforming all baselines by a clear margin (+7.7 Avg. ACC over Search-R1 on 7B).
The gains are especially pronounced on the internal AD-QA benchmark (+20.6 over Search-R1 on 7B), a real-world advertising QA dataset with complex multi-hop queries, confirming the practical value of path-centric rewards in industrial settings.
Notably, the improvements are consistent across model scales, with the 3B model achieving +7.9 Avg. ACC over Search-R1, demonstrating that path-centric rewards are effective even for smaller models.

\subsection{Ablation Study}

We conduct ablation studies to validate the contribution of each reward component in \method: format reward, path-centric reward, and outcome reward.

\subsubsection{Format Reward}

\begin{figure}[t]
    \centering
    \includegraphics[width=\columnwidth]{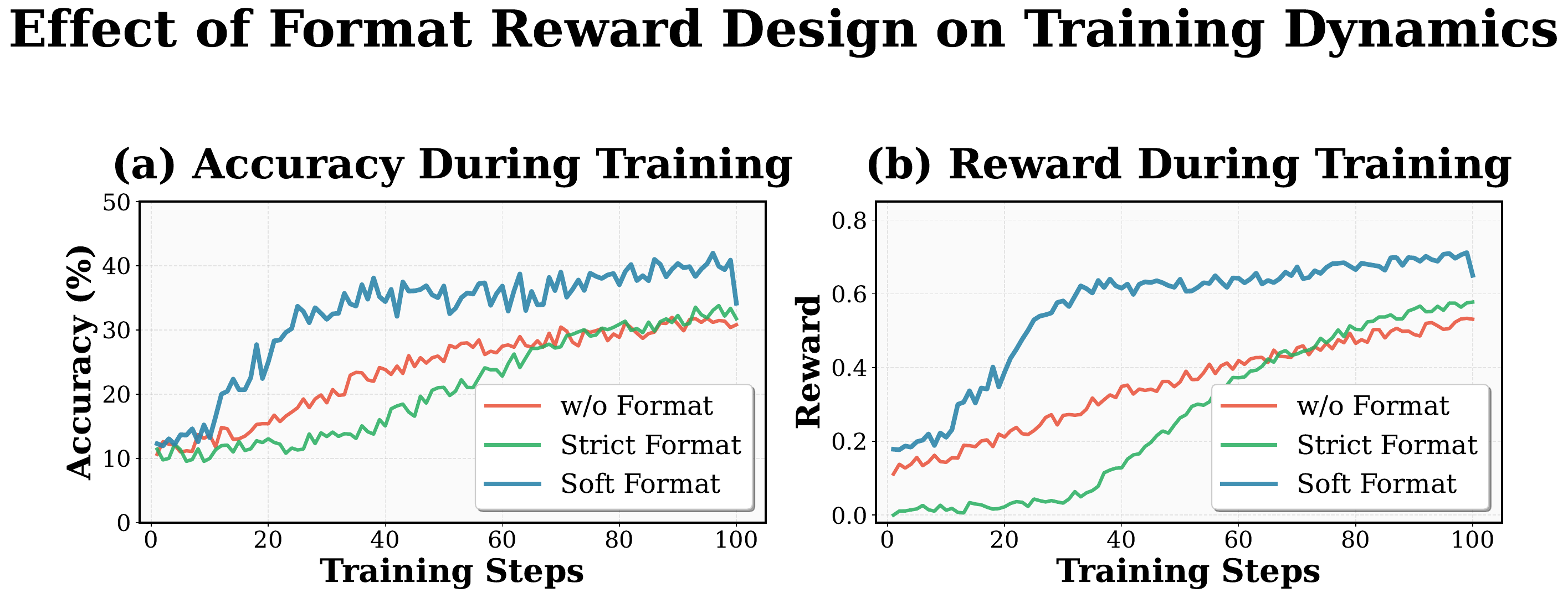}
    \caption{Training dynamics comparison of different format reward strategies. Soft Format (our buffered design) achieves faster ACC improvement and higher stable rewards compared to Strict Format (zero reward for invalid format) and Without Format baseline.}
    \label{fig:format_reward}
\end{figure}
\begin{figure}[t]
    \centering
    \includegraphics[width=\columnwidth]{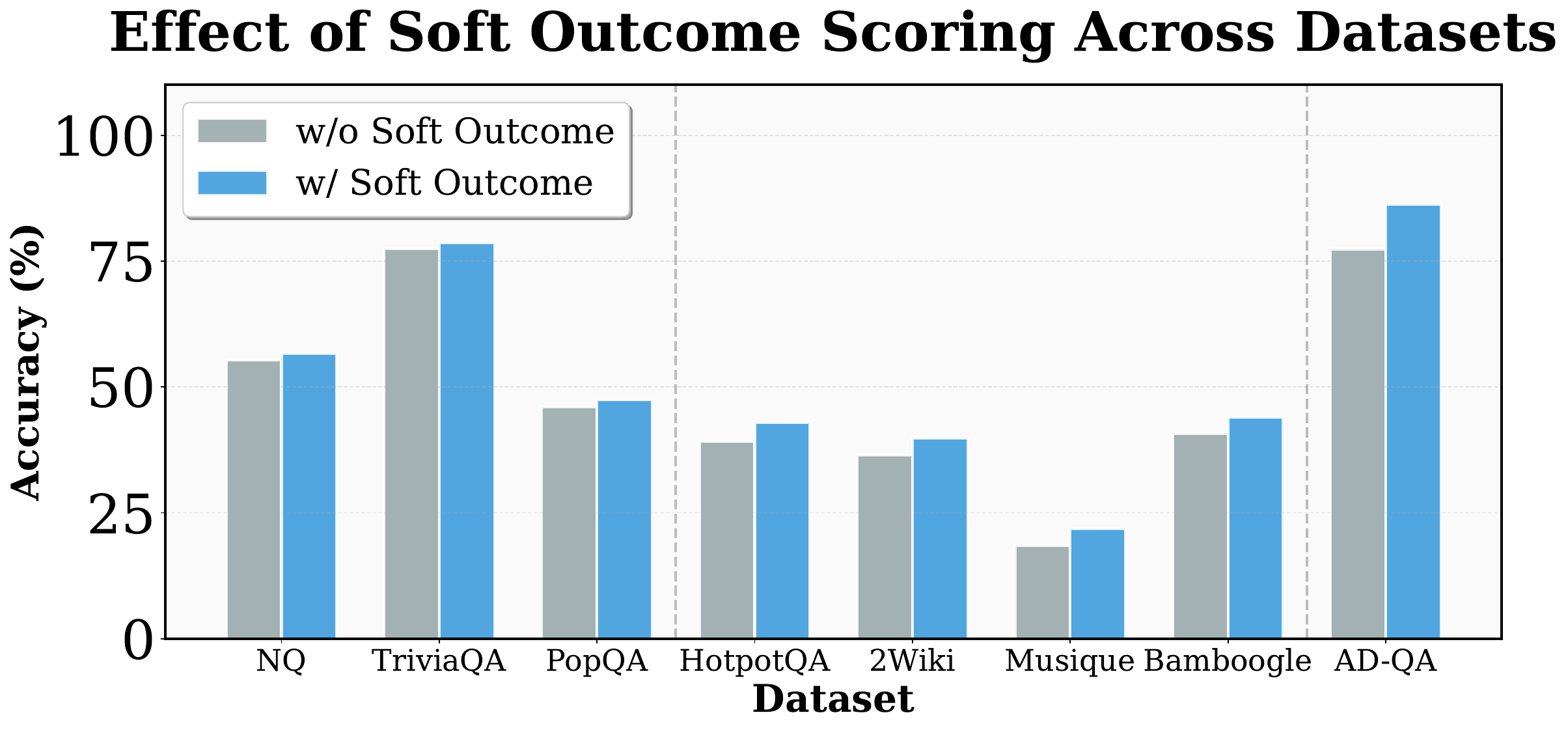}
\caption{Effect of soft outcome scoring across datasets. Gray bars show accuracy without soft scoring (binary outcome), blue bars show accuracy with soft scoring. Per-dataset results are in Appendix~\ref{sec:appendix_soft_outcome}.}
\label{fig:outcome_analysis}
\end{figure}
As shown in Figure~\ref{fig:format_reward}, we compare three strategies: (1) Soft Format (our buffered design), (2) Strict Format (zero reward for violations), and (3) Without Format.
Our soft format achieves significantly faster convergence by providing continuous gradient feedback, while the strict approach yields near-zero rewards in early training steps due to frequent formatting errors.

\subsubsection{Path-Centric Reward}
As shown in Table~\ref{tab:ablation}, removing reference-alignment causes a 5.3\% accuracy drop, confirming that reference planners provide valuable path-centric guidance.
Removing self-consistency results in a 3.1\% decrease. The full dual-track model achieves the best performance, validating that both external guidance and internal consistency are complementary signals.

\vspace{-1mm}
\subsubsection{Outcome Reward}
\vspace{-1mm}

As shown in Figure~\ref{fig:outcome_analysis}, soft outcome scoring provides modest gains for single-hop tasks (+1.2\%), larger improvements for multi-hop QA (+3.5\%), and the highest gain for AD-QA (+8.8\%), confirming that complex scenarios benefit most from partial credit signals.

\section{Analysis}
\label{sec:analysis}

\subsection{Hyperparameter Sensitivity}
\label{sec:hyperparameter}

We investigate the impact of two critical hyperparameters in our reward formulation: the path reward weight $\lambda_p$ and the accuracy weight $\lambda_a$.

\begin{figure}[t]
    \centering
    \includegraphics[width=\columnwidth]{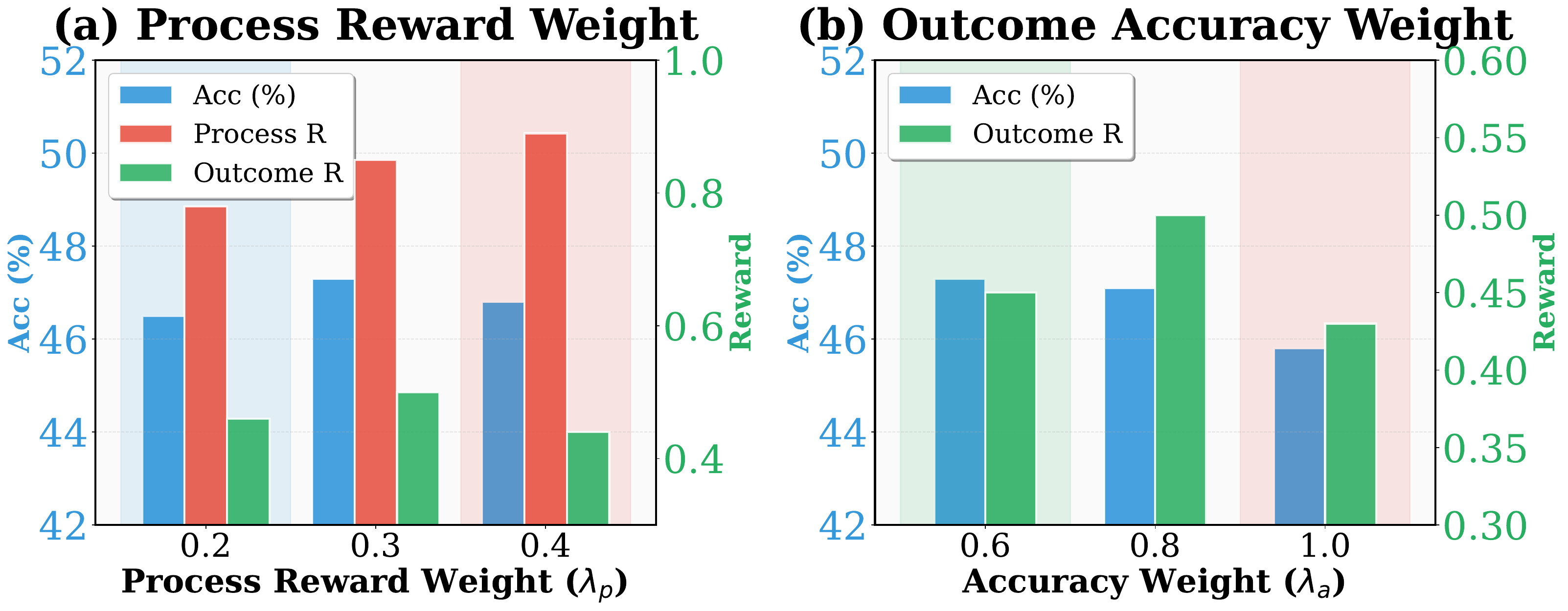}
\caption{Hyperparameter sensitivity analysis. All rewards are averaged over steps 195--205. (a) Effect of path reward weight $\lambda_p$. (b) Effect of accuracy weight $\lambda_a$. Per-dataset results are in Appendix~\ref{sec:appendix_hyperparameter}.}
\label{fig:hyperparameter}
\end{figure}

As shown in Figure~\ref{fig:hyperparameter}, both $\lambda_p$ and $\lambda_a$ exhibit clear sweet spots. Too little path weight provides insufficient supervision, while too much induces reward overfitting where path metrics improve but accuracy drops. Similarly, over-weighting accuracy neglects reasoning quality and leads to reward hacking. The optimal configuration ($\lambda_p{=}0.3$, $\lambda_a{=}0.6$) balances accuracy as the primary objective with reasoning quality as a regularizer.

\subsection{Efficiency Analysis}
\label{sec:turn_analysis}

\begin{figure}[t]
    \centering
    \includegraphics[width=\columnwidth]{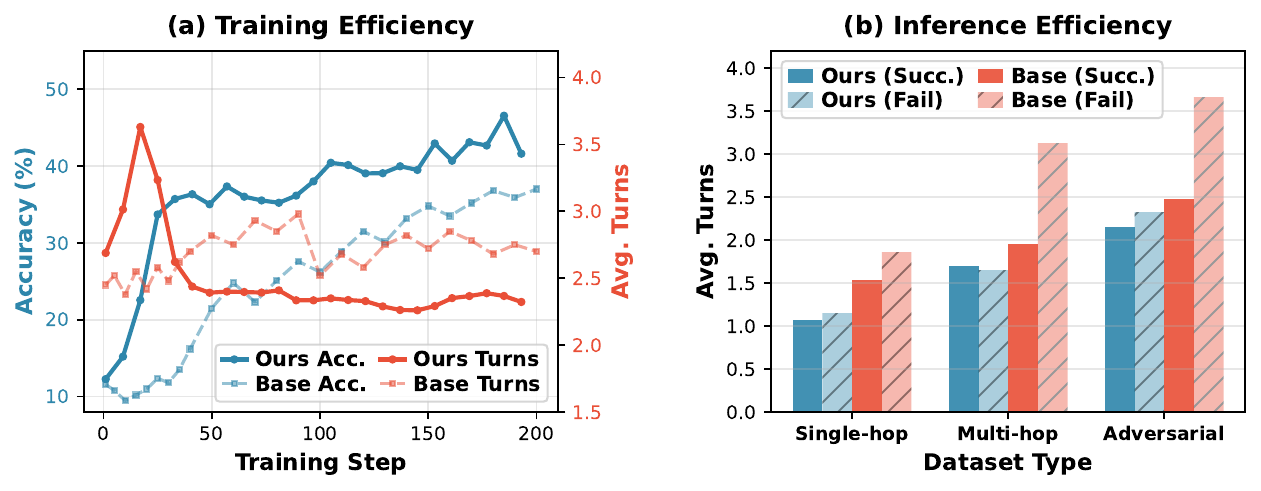}
    \caption{Efficiency analysis. (a) Training efficiency: accuracy and interaction turns comparison between \method and Search-R1 during training. (b) Inference efficiency: turns by outcome across dataset types.}
    \label{fig:efficiency_analysis}
\end{figure}

\paragraph{Training Efficiency}
Figure~\ref{fig:efficiency_analysis}(a) compares training dynamics. \method converges significantly faster, reaching Search-R1's final accuracy ($\sim$40\%) within 60 steps versus over 150. Meanwhile, \method's interaction turns steadily decrease, indicating path-centric rewards guide toward higher accuracy and more concise reasoning, while Search-R1's turns remain flat or increase.

\paragraph{Inference Efficiency}
Figure~\ref{fig:efficiency_analysis}(b) compares turn distributions across dataset types. Two key findings emerge: (1) Both methods require more turns for complex adversarial queries. (2) \method\ maintains consistent turn counts between successful and unsuccessful cases, while Search-R1 exhibits larger gaps for multi-hop (+60\%) and adversarial (+47\%) tasks.

\subsection{Model and RL Algorithm Analysis}
\label{sec:model_rl}

\begin{table}[t]
\centering
\small
\setlength{\tabcolsep}{5pt}
\begin{tabular}{ll|ccc}
\toprule
\textbf{Model} & \textbf{RL} & \textbf{Single} & \textbf{Multi} & \textbf{AD} \\
\midrule
Qwen2.5-3B & GRPO & \textbf{58.5} & \textbf{28.7} & \textbf{79.5} \\
Qwen2.5-3B & PPO & 57.2 & 27.5 & 77.8 \\
Llama-3.2-3B & GRPO & 56.8 & 27.1 & 76.3 \\
Llama-3.2-3B & PPO & 55.6 & 26.2 & 74.6 \\
\bottomrule
\end{tabular}
\caption{ACC (\%) across base models and RL algorithms. All models use Instruct versions. Per-dataset results are in Appendix~\ref{sec:appendix_model_rl}.}
\label{tab:model_rl}
\end{table}

Table~\ref{tab:model_rl} examines the impact of base models and RL algorithms.
Qwen2.5-3B-Instruct~\cite{qwen2025qwen25technicalreport} slightly outperforms Llama-3.2-3B-Instruct~\cite{grattafiori2024llama3herdmodels} across all task types, likely due to stronger instruction-following and reasoning capabilities in the base model.
GRPO~\cite{shao2024deepseekmathpushinglimitsmathematical} achieves marginally higher accuracy than PPO~\cite{schulman2017proximalpolicyoptimizationalgorithms}; however, PPO exhibits more stable training dynamics with lower variance across runs.
Importantly, path-centric rewards yield consistent gains across all model--algorithm combinations, suggesting that our approach is orthogonal to the choice of base model and RL algorithm.

\subsection{LLM Evaluator Analysis}
\label{sec:evaluator_analysis}

Our dual-track scoring and soft outcome scoring rely on an external LLM evaluator during training (at inference time, no evaluator calls are needed).
To examine sensitivity, we replaced the default evaluator (HY 2.0-Instruct) with Qwen3-32B and Qwen3-8B, and sampled 200 trajectories to measure human agreement. As shown in Table~\ref{tab:evaluator}, Qwen3-32B achieves comparable accuracy ($-$0.8) and human agreement, while Qwen3-8B degrades by 3.2 points with lower outcome scoring agreement (78.5\%). Nevertheless, step coverage---the core component of our path-centric reward---remains robust even with the 8B evaluator (88.0\% agreement), confirming that \method is not tightly coupled to a specific evaluator.

\begin{table}[t]
\centering
\small
\setlength{\tabcolsep}{5.5pt}
\begin{tabular}{l|c|ccc}
\toprule
\multirow{2}{*}{\textbf{Evaluator}} & \multirow{2}{*}{\textbf{ACC}} & \multicolumn{3}{c}{\textbf{Human Agree. (\%)}} \\
& & \textbf{Plan} & \textbf{Step} & \textbf{Outc.} \\
\midrule
HY 2.0-Inst. & 47.3 & 91.2 & 94.5 & 88.7 \\
Qwen3-32B & 46.5 & 89.0 & 92.5 & 85.0 \\
Qwen3-8B & 44.1 & 83.5 & 88.0 & 78.5 \\
\bottomrule
\end{tabular}
\caption{Effect of LLM evaluator choice on \method Avg. ACC and human agreement. Per-dataset results are in Appendix~\ref{sec:appendix_evaluator}.}
\label{tab:evaluator}
\end{table}

\subsection{Case Study}
\label{sec:case_study}

To qualitatively illustrate \method's advantages, we present case studies comparing reasoning trajectories with baseline methods. Appendix~\ref{sec:appendix_case} provides a representative example from multi-hop QA, demonstrating how path-centric rewards lead to more structured decomposition, precise query formulation, and effective information synthesis.

\section{Conclusion}
\label{sec:conclusion}

We presented \method, a framework that introduces path-centric reward shaping for agentic RAG training.
By evaluating the structural quality of entire reasoning paths rather than isolated elements, our approach provides fine-grained supervision while respecting the inherent diversity of multi-step reasoning.
Extensive experiments on public QA benchmarks and an internal advertsing dataset demonstrate significant improvements in accuracy and efficiency, validating path-centric rewards in both academic and industrial settings.

\clearpage

\section*{Ethics Statement}

Our work focuses on improving the training of AI systems for information retrieval and reasoning.
We use publicly available datasets for training and evaluation. The internal AD-QA dataset is fully anonymized with all personally identifiable information removed prior to use.
The improved efficiency of agentic RAG systems could reduce computational resources required for deployment, contributing to more sustainable AI.

\bibliography{custom}

\clearpage
\appendix

\section{AD-QA Dataset}
\label{sec:appendix_adqa}

AD-QA is a fully anonymized multi-hop QA benchmark from a real-world advertising domain, containing 1,000 test instances requiring multi-step reasoning across domains such as campaign configuration, bidding strategies, audience targeting, and conversion tracking. All instances are derived from authentic user queries with all personally identifiable information removed.

Each question requires synthesizing information from at least two distinct knowledge domains, making it a challenging benchmark for multi-hop reasoning in enterprise settings. Ground-truth answers are curated by domain experts and verified through cross-validation.

\section{Implementation Details}
\label{sec:appendix_impl}

\subsection{Training Configuration}

For GRPO training, we set the policy learning rate to $1 \times 10^{-6}$ with a warm-up ratio of 0.1. Training is conducted on 8$\times$H20 GPUs using a total batch size of 512, with a mini-batch size of 256. The micro-batch size per GPU is set to 8 for 7B models and 16 for 3B models.

The maximum prompt length and response length are both set to 4,096 tokens, with a maximum model context length of 8,192 tokens. We enable gradient checkpointing for memory efficiency and use Fully Sharded Data Parallel (FSDP) with reference model parameter offloading.

For efficient rollout generation, we use SGLang with tensor parallel size of 1 and GPU memory utilization of 0.8 (7B) or 0.75 (3B). Rollout sampling uses temperature $\tau = 0.6$, top-$k = 20$, and top-$p = 0.95$. We sample 16 candidate responses per prompt for 7B models and 32 for 3B models with an over-sample rate of 0.1. The KL divergence coefficient $\beta$ is set to 0.001 with low-variance KL loss, and the clip ratio ranges from 0.2 to 0.28.

\subsection{Reward Computation}

The path-centric reward combines three components with the following default weights: format reward weight $\lambda_f = 0.1$, path reward weight $\lambda_p = 0.3$, and outcome accuracy weight $\lambda_a = 0.6$. The reference planner uses a proprietary instruction-tuned model (anonymized as HY 2.0-Instruct) to generate guidance trajectories, which are cached offline before training to avoid runtime overhead.

For self-consistency scoring, we sample 3 independent reasoning paths per query and compute pairwise agreement using Jaccard similarity on extracted evidence spans. The soft outcome scoring applies a decay factor of 0.5 for partial matches when the final answer is incorrect but the reasoning path demonstrates high path quality.

\subsection{Computational Cost}
\label{sec:appendix_cost}

Reference planners are generated offline for all 90K training samples using HY 2.0-Instruct, with each sample requiring on average 1.91 LLM calls. This is a one-time cost cached before RL training and amortized over all subsequent runs.

\subsection{Inference Settings}

During inference, we set the maximum action budget $B = 4$, allowing up to 4 search-reason iterations per query. The retriever returns top-3 passages per search step. We use sampling with temperature 0.6 and top-$p$ 0.95 for validation. Model checkpoints are saved every 10 steps, and we select the checkpoint with the highest validation accuracy for final evaluation.

\section{Algorithms}
\label{sec:appendix_algo}

This section provides algorithmic descriptions of the key components in \method: (1) offline reference planner generation, (2) agentic RAG inference, and (3) path-centric reward computation.

Algorithm~\ref{alg:ref_planner} describes reference planner generation using a high-capability LLM (HY 2.0-Instruct) to produce structured plans and reference reasoning paths, cached offline for training.

Algorithm~\ref{alg:inference} illustrates agentic RAG inference: the model iteratively generates reasoning, issues search queries via \texttt{<tool\_call>}, and receives retrieved passages as \texttt{<tool\_response>} until the action budget is exhausted or an answer is produced.

Algorithm~\ref{alg:reward} details the reward computation combining format, dual-track path-centric, and soft outcome signals.

\begin{algorithm*}[t]
\caption{Reference Planner Generation}
\label{alg:ref_planner}
\begin{algorithmic}[1]
\REQUIRE Training dataset $\mathcal{D} = \{(q_i, a_i)\}_{i=1}^{N}$, reference LLM $\mathcal{M}_{\text{ref}}$
\ENSURE Reference trajectories $\mathcal{T}_{\text{ref}} = \{(p_i, r_i)\}_{i=1}^{N}$
\FOR{each $(q, a) \in \mathcal{D}$}
    \STATE $\text{prompt}_p \gets$ \textsc{PlannerPrompt}$(q)$ \COMMENT{Generate planning prompt}
    \STATE $p \gets \mathcal{M}_{\text{ref}}(\text{prompt}_p)$ \COMMENT{Generate reference plan}
    \STATE $\text{prompt}_r \gets$ \textsc{ReasoningPrompt}$(q, p)$ \COMMENT{Generate reasoning prompt}
    \STATE $r \gets \mathcal{M}_{\text{ref}}(\text{prompt}_r)$ \COMMENT{Generate reference reasoning path}
    \STATE $\mathcal{T}_{\text{ref}} \gets \mathcal{T}_{\text{ref}} \cup \{(p, r)\}$
\ENDFOR
\RETURN $\mathcal{T}_{\text{ref}}$
\end{algorithmic}
\end{algorithm*}

\begin{algorithm*}[t]
\caption{Agentic RAG Inference}
\label{alg:inference}
\begin{algorithmic}[1]
\REQUIRE Question $q$, policy model $\pi$, retriever $\mathcal{R}$, action budget $B$, top-$K$
\ENSURE Generated trajectory $y$ with final answer
\STATE $y \gets \texttt{<reasoning>}$; \quad $t \gets 1$
\WHILE{$t \le B$}
    \STATE $\Delta \gets \textsc{Generate}(\pi, y)$ \textbf{until} $\texttt{</tool\_call>}$ or $\texttt{</answer>}$
    \STATE $y \gets y \,\|\, \Delta$
    \IF{$\textsc{Contains}(y, \texttt{</answer>})$}
        \STATE \textbf{break} \COMMENT{Final answer generated}
    \ENDIF
    \IF{$\textsc{Contains}(\Delta, \texttt{<tool\_call>})$}
        \STATE $\text{query} \gets \textsc{Extract}(\Delta, \texttt{<tool\_call>})$
        \STATE $\text{docs} \gets \mathcal{R}(\text{query}, K)$ \COMMENT{Retrieve top-$K$ passages}
        \STATE $y \gets y \,\|\, \texttt{<tool\_response>} \,\|\, \text{docs} \,\|\, \texttt{</tool\_response>}$
        \STATE $t \gets t + 1$
    \ENDIF
\ENDWHILE
\IF{\textbf{not} $\textsc{Contains}(y, \texttt{</answer>})$}
    \STATE $y \gets y \,\|\, \texttt{<answer>} \,\|\, \textsc{Generate}(\pi, y)$ \textbf{until} $\texttt{</answer>}$
\ENDIF
\RETURN $y$
\end{algorithmic}
\end{algorithm*}

\begin{algorithm*}[t]
\caption{\method Reward Computation}
\label{alg:reward}
\begin{algorithmic}[1]
\REQUIRE Trajectory $y$, ground truth $a^*$, reference plan $p_{\text{ref}}$, reference path $r_{\text{ref}}$
\ENSURE Total reward $R(y)$
\STATE \textcolor{gray}{\textit{// Format Reward}}
\IF{$\textsc{ValidFormat}(y)$ \textbf{and} $\textsc{HasAnswer}(y)$ \textbf{and} $\textsc{HasToolCall}(y)$}
    \STATE $r_f \gets 0.1$
\ELSIF{$\textsc{HasAnswer}(y)$ \textbf{and} $\textsc{HasToolResponse}(y)$}
    \STATE $r_f \gets 0.05$
\ELSE
    \RETURN $0$ \COMMENT{Invalid trajectory}
\ENDIF
\STATE
\STATE \textcolor{gray}{\textit{// Path-Centric Reward via Dual-Track Evaluation}}
\STATE $\text{eval} \gets \textsc{LLMEvaluate}(y, p_{\text{ref}}, r_{\text{ref}})$ \COMMENT{Call evaluator LLM}
\STATE $r_{\text{planner}} \gets \text{eval}.\text{planner\_score}$ \COMMENT{Plan quality: 0.2/0.6/1.0/1.2}
\STATE
\STATE \textcolor{gray}{\textit{// Track A: Self-Consistency}}
\STATE $s_{\text{self}} \gets r_{\text{planner}} \times \frac{\text{eval}.\text{eff\_steps\_self}}{\text{eval}.\text{model\_plan\_steps}}$
\STATE
\STATE \textcolor{gray}{\textit{// Track B: Reference-Alignment}}
\STATE $s_{\text{ref}} \gets \frac{\text{eval}.\text{eff\_steps\_ref}}{|\text{steps}(r_{\text{ref}})|}$
\STATE
\STATE $r_p \gets \max(s_{\text{self}}, s_{\text{ref}})$ \COMMENT{Best of dual tracks}
\STATE
\STATE \textcolor{gray}{\textit{// Outcome Reward with Soft Scoring}}
\IF{$\textsc{ExactMatch}(\textsc{GetAnswer}(y), a^*)$}
    \STATE $r_o \gets 1.0$
\ELSE
    \STATE $r_o \gets 0.8 \times \text{eval}.\text{acc\_score} + 0.2 \times \text{eval}.\text{reason\_score}$
\ENDIF
\STATE
\STATE $R(y) \gets \lambda_f \cdot r_f + \lambda_p \cdot r_p + \lambda_o \cdot r_o$
\RETURN $R(y)$
\end{algorithmic}
\end{algorithm*}

\section{Prompt Templates}
\label{sec:appendix_prompts}

This section presents the prompt templates used in \method for inference, reference planner generation, and reward evaluation.

\subsection{Agentic RAG Inference Prompt}

Figure~\ref{fig:prompt_inference} shows the prompt template used during both training rollouts and inference. The prompt instructs the model to decompose questions into sub-tasks, execute searches iteratively, and produce structured outputs with \texttt{<reasoning>}, \texttt{<tool\_call>}, and \texttt{<answer>} tags.

\begin{figure*}[t]
\centering
\begin{tcolorbox}[colback=gray!5, colframe=gray!50, title=\textbf{Agentic RAG Inference Prompt}, fonttitle=\bfseries, boxrule=0.5pt, left=3mm, right=3mm, top=2mm, bottom=2mm]
You are a meticulous \textbf{Deep Research Agent}. Your goal is to provide a comprehensive and accurate answer by conducting multiple rounds of search.

\textbf{\#\# CRITICAL INSTRUCTIONS}

\textbf{1. Detailed Planning (<reasoning>):}
\begin{itemize}[leftmargin=*, nosep]
\item In the first turn, you MUST break the question down into \textbf{multiple dependent sub-questions}.
\item Focus on one sub-question at a time.
\end{itemize}

\textbf{2. Step-by-Step Execution (<tool\_call>):}
\begin{itemize}[leftmargin=*, nosep]
\item Execute only ONE search query per turn.
\item After receiving results, verify: ``Is this sufficient? Do I need more details?''
\end{itemize}

\textbf{3. No Guessing:}
\begin{itemize}[leftmargin=*, nosep]
\item If results are incomplete, issue another search. Do NOT hallucinate.
\end{itemize}

\textbf{4. Final Answer (<answer>):}
\begin{itemize}[leftmargin=*, nosep]
\item Only output <answer> when ALL necessary information is gathered.
\end{itemize}

\textbf{\#\# CURRENT TASK}

Question: \texttt{\{question\}}
\end{tcolorbox}
\caption{Prompt template for agentic RAG inference. The model is instructed to plan, search iteratively, and provide structured outputs.}
\label{fig:prompt_inference}
\end{figure*}

\subsection{Reference Planner Generation Prompt}

Figure~\ref{fig:prompt_planner} shows the prompt used to generate reference plans and reasoning paths from HY 2.0-Instruct. Given a question and its correct answer, the reference LLM produces an optimized search strategy that serves as guidance during path reward computation.

\begin{figure*}[t]
\centering
\begin{tcolorbox}[colback=blue!3, colframe=blue!40, title=\textbf{Reference Planner Generation Prompt}, fonttitle=\bfseries, boxrule=0.5pt, left=3mm, right=3mm, top=2mm, bottom=2mm]
You are an expert planner and reasoning optimizer.

\textbf{Current Question:} \texttt{\{question\}}

\textbf{Correct Answer:} \texttt{\{golden\_answers\}}

Your task is to generate:

\textbf{1. Optimized Reasoning Path:} A sequence of search queries that would lead directly to the correct answer in the most efficient way. Format as a numbered list.

\textbf{2. Optimized Planner:} A concise, step-by-step instruction on how a reasoning agent should solve this question correctly and efficiently.

\textbf{Important:}
\begin{itemize}[leftmargin=*, nosep]
\item Focus on the minimal set of queries needed.
\item Avoid redundant or inefficient steps.
\end{itemize}

\textbf{Output format:}

\texttt{<correct\_reasoning\_path>}\\
\texttt{1. query 1}\\
\texttt{2. query 2}\\
\texttt{</correct\_reasoning\_path>}

\texttt{<optimized\_planner>}\\
\texttt{To solve this, first search for... then...}\\
\texttt{</optimized\_planner>}
\end{tcolorbox}
\caption{Prompt template for reference planner generation. HY 2.0-Instruct generates optimal search strategies for each training sample.}
\label{fig:prompt_planner}
\end{figure*}

\subsection{Dual-Track Evaluation Prompt}
\label{sec:appendix_eval_prompt}

Figure~\ref{fig:prompt_eval} presents the prompt used for dual-track path evaluation. An evaluator LLM assesses the model's trajectory along two dimensions: self-consistency (execution of its own plan) and reference-alignment (coverage of expert reference steps), along with outcome quality scoring.

\begin{figure*}[t]
\centering
\begin{tcolorbox}[colback=green!3, colframe=green!40, title=\textbf{Dual-Track Evaluation Prompt}, fonttitle=\bfseries, boxrule=0.5pt, left=3mm, right=3mm, top=2mm, bottom=2mm]
You are an expert RL researcher evaluating an AI agent's trajectory.

Your task is to conduct a \textbf{Dual-Track Evaluation}:
\begin{enumerate}[leftmargin=*, nosep]
\item \textbf{Self-Consistency Track}: How well did the agent execute its OWN plan?
\item \textbf{Reference-Alignment Track}: How well did the agent follow the Expert plan?
\item \textbf{Outcome Evaluation}: Assess accuracy and reasoning quality.
\end{enumerate}

\textbf{Evaluation Inputs:}
\begin{itemize}[leftmargin=*, nosep]
\item \textbf{Question}: \texttt{\{question\}}
\item \textbf{Correct Answer}: \texttt{\{golden\_answers\}}
\item \textbf{Reference Planner}: \texttt{\{ref\_planner\}}
\item \textbf{Reference Path}: \texttt{\{ref\_reasoning\_path\}}
\item \textbf{Model Trajectory}: \texttt{\{trajectory\}}
\end{itemize}

\textbf{Scoring Criteria:}
\begin{itemize}[leftmargin=*, nosep]
\item \textbf{Planner Score}: 0.2 (Bad) / 0.6 (Average) / 1.0 (Good) / 1.2 (Excellent)
\item \textbf{Outcome Accuracy}: 0.0 (Wrong) / 0.5 (Partial) / 1.0 (Correct)
\item \textbf{Reasoning Quality}: 0.0 / 0.5 / 0.8 / 1.0
\end{itemize}

\textbf{Output:} JSON with \texttt{planner\_score}, \texttt{model\_plan\_steps}, \texttt{effective\_steps\_self}, \texttt{effective\_steps\_ref}, \texttt{outcome\_accuracy\_score}, \texttt{outcome\_reasoning\_score}.
\end{tcolorbox}
\caption{Prompt template for dual-track evaluation. The evaluator LLM assesses both self-consistency and reference-alignment of model trajectories.}
\label{fig:prompt_eval}
\end{figure*}

\section{Case Study}
\label{sec:appendix_case}

To qualitatively illustrate \method's advantages, we present a representative case from MuSiQue demonstrating how path-centric reward shaping leads to more accurate multi-hop reasoning.

\subsection{Multi-Hop Reasoning Comparison}

Figure~\ref{fig:case_comparison} compares Search-R1 and \method on a multi-hop question. Without explicit planning, Search-R1 misinterprets ``rock \& roll'' as a genre descriptor, retrieving information about the wrong entity. In contrast, \method's planning correctly identifies ``Bang Bang Rock \& Roll'' as a complete album title, leading to the correct answer.

\begin{figure*}[t]
\centering
\begin{tcolorbox}[colback=gray!5, colframe=gray!60, title=\textbf{Query: Who is the lead singer of the band performing Bang Bang Rock \& Roll? \hfill Answer: Eddie Argos}, fonttitle=\bfseries, boxrule=0.8pt, left=2mm, right=2mm, top=2mm, bottom=2mm]

\begin{minipage}[t]{0.48\textwidth}
\begin{tcolorbox}[colback=red!3, colframe=red!40, title=\textbf{Search-R1 (Incorrect)}, fonttitle=\bfseries, boxrule=0.5pt, left=2mm, right=2mm, top=1mm, bottom=1mm]
\texttt{<think>} I need to find who performs ``Bang Bang Rock \& Roll.'' \texttt{</think>}

\colorbox{yellow!30}{\texttt{<search>}} \textcolor{red}{\textit{song Bang Bang rock and roll genre}} \colorbox{yellow!30}{\texttt{</search>}}

\texttt{<result>} In 2017, Dan Auerbach collaborated with Robert Finley on ``Bang Bang,'' featuring a classic rock and roll sound... \texttt{</result>}

\colorbox{yellow!30}{\texttt{<search>}} \textit{lead singer of The Black Keys} \colorbox{yellow!30}{\texttt{</search>}}

\texttt{<result>} Dan Auerbach serves as the lead vocalist... \texttt{</result>}

\textbf{Final:} \textcolor{red}{\ding{55}} \textcolor{red}{\textbf{Dan Auerbach}}
\end{tcolorbox}
\end{minipage}
\hfill
\begin{minipage}[t]{0.48\textwidth}
\begin{tcolorbox}[colback=green!3, colframe=green!40, title=\textbf{\method (Correct)}, fonttitle=\bfseries, boxrule=0.5pt, left=2mm, right=2mm, top=1mm, bottom=1mm]
\texttt{<reasoning>} I need to: 1. Find the band performing ``Bang Bang Rock \& Roll''. 2. Find the lead singer. \texttt{</reasoning>}

\colorbox{yellow!30}{\texttt{<tool\_call>}} \textcolor{teal}{\textit{``Bang Bang Rock \& Roll'' album}} \colorbox{yellow!30}{\texttt{</tool\_call>}}

\texttt{<tool\_response>} ``Bang Bang Rock \& Roll'' is the debut album by British rock band \textcolor{teal}{\textbf{Art Brut}}... \texttt{</tool\_response>}

\colorbox{yellow!30}{\texttt{<tool\_call>}} \textit{lead singer of Art Brut} \colorbox{yellow!30}{\texttt{</tool\_call>}}

\texttt{<tool\_response>} ...frontman \textcolor{teal}{\textbf{Eddie Argos}}' enthusiastic vocal delivery... \texttt{</tool\_response>}

\textbf{Final:} \textcolor{green!50!black}{\ding{51}} \textcolor{teal}{\textbf{Eddie Argos}}
\end{tcolorbox}
\end{minipage}

\end{tcolorbox}
\caption{Comparison of reasoning trajectories. Search-R1's imprecise query retrieves valid but irrelevant results; \method's planning-driven query retrieves the correct information. \colorbox{yellow!30}{Highlighted} text shows search queries.}
\label{fig:case_comparison}
\end{figure*}

\section{Additional Results}
\label{sec:appendix_results}

\subsection{Impact of Retrieved Documents per Search}
\label{sec:appendix_num_docs}

\begin{table*}[h]
\centering
\small
\begin{tabular}{cl|ccc|cccc|c|c}
\toprule
\multirow{2}{*}{\textbf{Model}} & \multirow{2}{*}{\textbf{\# Docs}} & \multicolumn{3}{c|}{\textbf{General QA}} & \multicolumn{4}{c|}{\textbf{Multi-Hop QA}} & \multirow{2}{*}{\textbf{Avg.}} & \textbf{Internal} \\
& & \textbf{NQ} & \textbf{TriviaQA} & \textbf{PopQA} & \textbf{HotpotQA} & \textbf{2Wiki} & \textbf{MuSiQue} & \textbf{Bamboogle} & & \textbf{AD-QA} \\
\midrule
\multirow{5}{*}{\textit{7B}} 
& 1 & 45.8 & 67.2 & 38.5 & 35.2 & 32.0 & 15.5 & 36.0 & 38.6 & 72.8 \\
& 2 & 52.0 & 74.2 & 44.0 & 39.5 & 36.8 & 18.8 & 41.0 & 43.8 & 80.5 \\
& 3 & \textbf{56.6} & \textbf{78.6} & \textbf{47.5} & \textbf{42.9} & \textbf{39.8} & 21.8 & 44.0 & \textbf{47.3} & \textbf{86.2} \\
& 5 & 55.8 & 77.8 & 46.8 & 42.5 & 39.2 & \textbf{22.4} & \textbf{45.2} & 47.1 & 85.5 \\
& 10 & 53.5 & 74.5 & 44.5 & 40.5 & 37.2 & 20.1 & 42.8 & 44.7 & 82.5 \\
\midrule
\multirow{5}{*}{\textit{3B}} 
& 1 & 42.8 & 63.5 & 38.2 & 28.8 & 28.5 & 9.2 & 21.2 & 33.2 & 64.5 \\
& 2 & 48.5 & 69.8 & 43.6 & 32.8 & 33.2 & 11.5 & 25.2 & 37.8 & 73.2 \\
& 3 & \textbf{53.0} & \textbf{74.5} & \textbf{47.9} & 36.2 & \textbf{36.6} & \textbf{13.3} & 28.8 & \textbf{41.5} & \textbf{79.5} \\
& 5 & 52.2 & 73.8 & 47.2 & \textbf{36.5} & 36.0 & 12.8 & \textbf{30.4} & 41.3 & 78.8 \\
& 10 & 49.8 & 71.2 & 45.0 & 33.8 & 34.2 & 11.8 & 27.6 & 39.1 & 75.5 \\
\bottomrule
\end{tabular}
\caption{Performance (ACC \%) with different numbers of retrieved documents per search. Retrieving 3 documents achieves the best average performance. While 5 documents shows advantages on specific datasets (MuSiQue, Bamboogle for 7B; HotpotQA, Bamboogle for 3B), the overall best configuration is 3 documents.}
\label{tab:num_docs}
\end{table*}

Table~\ref{tab:num_docs} shows how the number of retrieved documents per search iteration affects model performance. Retrieving too few documents may miss relevant information, while retrieving too many can introduce noise and increase context length.

\subsection{Effect of Format Reward on Output Compliance}
\label{sec:appendix_format}

\begin{table*}[h]
\centering
\small
\begin{tabular}{cl|ccc|cccc|c}
\toprule
\multirow{2}{*}{\textbf{Model}} & \multirow{2}{*}{\textbf{Method}} & \multicolumn{3}{c|}{\textbf{General QA}} & \multicolumn{4}{c|}{\textbf{Multi-Hop QA}} & \textbf{Internal} \\
& & \textbf{NQ} & \textbf{TriviaQA} & \textbf{PopQA} & \textbf{HotpotQA} & \textbf{2Wiki} & \textbf{MuSiQue} & \textbf{Bamboogle} & \textbf{AD-QA} \\
\midrule
\multirow{2}{*}{\textit{7B}} 
& w/o Format Reward & 82.8 & 84.2 & 81.5 & 79.1 & 76.8 & 75.2 & 83.6 & 88.2 \\
& w/ Format Reward & \textbf{95.1} & \textbf{95.6} & \textbf{94.2} & \textbf{91.5} & \textbf{88.8} & \textbf{87.6} & \textbf{96.4} & \textbf{97.5} \\
\midrule
\multirow{2}{*}{\textit{3B}} 
& w/o Format Reward & 72.8 & 75.5 & 72.2 & 67.1 & 61.8 & 62.4 & 65.2 & 78.8 \\
& w/ Format Reward & \textbf{87.2} & \textbf{88.1} & \textbf{85.4} & \textbf{80.2} & \textbf{74.8} & \textbf{75.1} & \textbf{78.0} & \textbf{91.5} \\
\bottomrule
\end{tabular}
\caption{Format compliance rate (\%) with and without format reward. Adding format reward significantly improves the model's ability to produce properly structured responses with parseable answers.}
\label{tab:format_compliance}
\end{table*}

Table~\ref{tab:format_compliance} analyzes the relationship between the format reward component and the model's ability to produce properly formatted outputs.

\subsection{Search Iterations Analysis}
\label{sec:appendix_iterations}

\begin{table*}[h]
\centering
\small
\begin{tabular}{cl|ccc|ccc}
\toprule
\multirow{2}{*}{\textbf{Model}} & \multirow{2}{*}{\textbf{Dataset}} & \multicolumn{3}{c|}{\textbf{Successful Cases}} & \multicolumn{3}{c}{\textbf{Failed Cases}} \\
\cmidrule(lr){3-5} \cmidrule(lr){6-8}
& & \textbf{1 iter} & \textbf{2 iter} & \textbf{3+ iter} & \textbf{1 iter} & \textbf{2 iter} & \textbf{3+ iter} \\
\midrule
\multirow{9}{*}{\textit{7B}} 
& NQ & 68.5\% & 22.3\% & 9.2\% & 45.2\% & 28.6\% & 26.2\% \\
& TriviaQA & 72.1\% & 19.8\% & 8.1\% & 48.3\% & 26.4\% & 25.3\% \\
& PopQA & 65.8\% & 24.5\% & 9.7\% & 42.1\% & 29.8\% & 28.1\% \\
& HotpotQA & 28.5\% & 48.2\% & 23.3\% & 35.4\% & 30.1\% & 34.5\% \\
& 2Wiki & 25.1\% & 50.6\% & 24.3\% & 33.8\% & 29.5\% & 36.7\% \\
& MuSiQue & 18.5\% & 52.3\% & 29.2\% & 31.2\% & 28.6\% & 40.2\% \\
& Bamboogle & 22.4\% & 45.6\% & 32.0\% & 28.3\% & 25.4\% & 46.3\% \\
& AD-QA & 35.2\% & 42.5\% & 22.3\% & 22.8\% & 28.5\% & 48.7\% \\
& \textbf{Average} & 42.0\% & 38.2\% & 19.8\% & 35.9\% & 28.4\% & 35.7\% \\
\midrule
\multirow{9}{*}{\textit{3B}} 
& NQ & 62.3\% & 26.1\% & 11.6\% & 40.5\% & 30.2\% & 29.3\% \\
& TriviaQA & 66.8\% & 23.4\% & 9.8\% & 44.1\% & 28.5\% & 27.4\% \\
& PopQA & 60.2\% & 28.3\% & 11.5\% & 38.6\% & 31.2\% & 30.2\% \\
& HotpotQA & 22.4\% & 45.8\% & 31.8\% & 30.2\% & 28.5\% & 41.3\% \\
& 2Wiki & 20.3\% & 47.2\% & 32.5\% & 28.5\% & 27.8\% & 43.7\% \\
& MuSiQue & 14.2\% & 48.6\% & 37.2\% & 26.4\% & 26.2\% & 47.4\% \\
& Bamboogle & 16.8\% & 42.1\% & 41.1\% & 24.5\% & 23.8\% & 51.7\% \\
& AD-QA & 28.5\% & 40.2\% & 31.3\% & 18.2\% & 25.6\% & 56.2\% \\
& \textbf{Average} & 36.4\% & 37.7\% & 25.9\% & 31.4\% & 27.7\% & 40.9\% \\
\bottomrule
\end{tabular}
\caption{Distribution of search iterations for successful and failed cases. General QA datasets (NQ, TriviaQA, PopQA) show high success rates with single-iteration searches, while Multi-Hop QA datasets require more iterations. Failed cases consistently show higher proportions of 3+ iterations, suggesting that excessive searching indicates difficulty in finding relevant information.}
\label{tab:search_iterations}
\end{table*}

Table~\ref{tab:search_iterations} presents the distribution of search iterations for successful and failed cases across different datasets.

\paragraph{Key Observations.} (1) General QA datasets achieve most successes with single-iteration searches. (2) Multi-hop datasets show successful cases concentrated at 2 iterations. (3) Failed cases consistently show higher 3+ iteration rates, suggesting excessive searching indicates difficulty. (4) The 3B model requires slightly more iterations than 7B.

\subsection{Detailed Ablation on Path-Centric Reward Components}
\label{sec:appendix_ablation_process}

\begin{table*}[h]
\centering
\small
\begin{tabular}{l|ccc|cccc|c|c}
\toprule
\multirow{2}{*}{\textbf{Method}} & \multicolumn{3}{c|}{\textbf{General QA}} & \multicolumn{4}{c|}{\textbf{Multi-Hop QA}} & \multirow{2}{*}{\textbf{Avg.}} & \textbf{Internal} \\
& \textbf{NQ} & \textbf{TriviaQA} & \textbf{PopQA} & \textbf{HotpotQA} & \textbf{2Wiki} & \textbf{MuSiQue} & \textbf{Bamboogle} & & \textbf{AD-QA} \\
\midrule
\multicolumn{10}{c}{\textit{Qwen2.5-7B}} \\
\midrule
\method (Full) & \textbf{56.6} & \textbf{78.6} & \textbf{47.5} & \textbf{42.9} & \textbf{39.8} & \textbf{21.8} & \textbf{44.0} & \textbf{47.3} & \textbf{86.2} \\
\quad w/o Reference-Alignment & 50.8 & 73.0 & 43.2 & 37.5 & 34.0 & 16.8 & 39.5 & 42.1 & 78.8 \\
\quad w/o Self-Consistency & 53.0 & 75.5 & 44.8 & 39.8 & 37.2 & 19.2 & 40.2 & 44.2 & 82.5 \\
\midrule
Search-R1 (Baseline) & 42.9 & 62.3 & 42.7 & 38.6 & 34.6 & 16.2 & 40.0 & 39.6 & 65.6 \\
\midrule
\multicolumn{10}{c}{\textit{Qwen2.5-3B}} \\
\midrule
\method (Full) & \textbf{53.0} & \textbf{74.5} & \textbf{47.9} & \textbf{36.2} & \textbf{36.6} & \textbf{13.3} & \textbf{28.8} & \textbf{41.5} & \textbf{79.5} \\
\quad w/o Reference-Alignment & 47.8 & 68.5 & 43.4 & 31.2 & 30.5 & 10.1 & 24.0 & 36.5 & 70.5 \\
\quad w/o Self-Consistency & 49.8 & 71.8 & 45.2 & 33.8 & 34.0 & 11.8 & 26.0 & 38.9 & 75.2 \\
\midrule
Search-R1 (Baseline) & 39.7 & 56.5 & 39.1 & 33.1 & 31.0 & 12.4 & 23.2 & 33.6 & 58.3 \\
\bottomrule
\end{tabular}
\caption{Detailed ablation study on path reward components (ACC \%). Removing reference-alignment causes larger drops on multi-hop datasets where external guidance is more critical, while removing self-consistency affects general QA more where the model's own planning suffices.}
\label{tab:ablation_detailed}
\end{table*}

Table~\ref{tab:ablation_detailed} provides the complete per-dataset breakdown for the path-centric reward component ablation study (extending Table~\ref{tab:ablation} in the main paper).

\subsection{Detailed Model and RL Algorithm Analysis}
\label{sec:appendix_model_rl}

\begin{table*}[h]
\centering
\small
\begin{tabular}{ll|ccc|cccc|c|c}
\toprule
\multirow{2}{*}{\textbf{Model}} & \multirow{2}{*}{\textbf{RL}} & \multicolumn{3}{c|}{\textbf{General QA}} & \multicolumn{4}{c|}{\textbf{Multi-Hop QA}} & \multirow{2}{*}{\textbf{Avg.}} & \textbf{Internal} \\
& & \textbf{NQ} & \textbf{TriviaQA} & \textbf{PopQA} & \textbf{HotpotQA} & \textbf{2Wiki} & \textbf{MuSiQue} & \textbf{Bamboogle} & & \textbf{AD-QA} \\
\midrule
Qwen2.5-3B-Inst. & GRPO & \textbf{53.0} & \textbf{74.5} & \textbf{47.9} & \textbf{36.2} & \textbf{36.6} & \textbf{13.3} & \textbf{28.8} & \textbf{41.5} & \textbf{79.5} \\
Qwen2.5-3B-Inst. & PPO & 51.6 & 73.1 & 46.8 & 34.8 & 35.4 & 12.6 & 27.6 & 40.3 & 78.1 \\
Llama-3.2-3B-Inst. & GRPO & 50.2 & 71.8 & 45.4 & 33.9 & 34.1 & 11.8 & 26.0 & 39.0 & 75.8 \\
Llama-3.2-3B-Inst. & PPO & 48.9 & 70.2 & 44.2 & 32.8 & 33.0 & 10.9 & 24.8 & 37.8 & 74.2 \\
\bottomrule
\end{tabular}
\caption{Detailed accuracy (\%) across different base models and RL algorithms on all datasets. Qwen2.5 consistently outperforms Llama-3.2, and GRPO achieves slightly higher accuracy than PPO across all datasets.}
\label{tab:model_rl_detailed}
\end{table*}

Table~\ref{tab:model_rl_detailed} extends Table~\ref{tab:model_rl} with per-dataset accuracy for different base models and RL algorithms.

\subsection{Detailed Soft Outcome Scoring Analysis}
\label{sec:appendix_soft_outcome}

\begin{table*}[h]
\centering
\small
\begin{tabular}{cl|ccc|cccc|c|c}
\toprule
\multirow{2}{*}{\textbf{Model}} & \multirow{2}{*}{\textbf{Method}} & \multicolumn{3}{c|}{\textbf{General QA}} & \multicolumn{4}{c|}{\textbf{Multi-Hop QA}} & \multirow{2}{*}{\textbf{Avg.}} & \textbf{Internal} \\
& & \textbf{NQ} & \textbf{TriviaQA} & \textbf{PopQA} & \textbf{HotpotQA} & \textbf{2Wiki} & \textbf{MuSiQue} & \textbf{Bamboogle} & & \textbf{AD-QA} \\
\midrule
\multirow{3}{*}{\textit{7B}} 
& w/o Soft Scoring & 55.4 & 77.5 & 46.0 & 39.2 & 36.5 & 18.5 & 40.8 & 44.8 & 77.4 \\
& w/ Soft Scoring & \textbf{56.6} & \textbf{78.6} & \textbf{47.5} & \textbf{42.9} & \textbf{39.8} & \textbf{21.8} & \textbf{44.0} & \textbf{47.3} & \textbf{86.2} \\
& \quad \textit{$\Delta$} & \textit{+1.2} & \textit{+1.1} & \textit{+1.5} & \textit{+3.7} & \textit{+3.3} & \textit{+3.3} & \textit{+3.2} & \textit{+2.5} & \textit{+8.8} \\
\midrule
\multirow{3}{*}{\textit{3B}} 
& w/o Soft Scoring & 51.8 & 73.2 & 46.5 & 32.6 & 33.1 & 10.4 & 24.5 & 38.9 & 68.5 \\
& w/ Soft Scoring & \textbf{53.0} & \textbf{74.5} & \textbf{47.9} & \textbf{36.2} & \textbf{36.6} & \textbf{13.3} & \textbf{28.8} & \textbf{41.5} & \textbf{79.5} \\
& \quad \textit{$\Delta$} & \textit{+1.2} & \textit{+1.3} & \textit{+1.4} & \textit{+3.6} & \textit{+3.5} & \textit{+2.9} & \textit{+4.3} & \textit{+2.6} & \textit{+11.0} \\
\bottomrule
\end{tabular}
\caption{Effect of soft outcome scoring (ACC \%). Multi-hop QA datasets benefit more from soft scoring (+3.0--3.7\%) compared to general QA datasets (+1.1--1.5\%), while the internal AD-QA dataset shows the largest improvement (+8.8--11.0\%), confirming that complex enterprise queries benefit most from partial credit signals.}
\label{tab:soft_outcome_detailed}
\end{table*}

Table~\ref{tab:soft_outcome_detailed} provides the per-dataset breakdown of the soft outcome scoring ablation (corresponding to Figure~\ref{fig:outcome_analysis} in the main paper).

\subsection{Detailed Hyperparameter Sensitivity Analysis}
\label{sec:appendix_hyperparameter}

\begin{table*}[h]
\centering
\small
\begin{tabular}{c|ccc|cccc|c|c}
\toprule
\multirow{2}{*}{\textbf{$\lambda_p$}} & \multicolumn{3}{c|}{\textbf{General QA}} & \multicolumn{4}{c|}{\textbf{Multi-Hop QA}} & \multirow{2}{*}{\textbf{Avg.}} & \textbf{Internal} \\
& \textbf{NQ} & \textbf{TriviaQA} & \textbf{PopQA} & \textbf{HotpotQA} & \textbf{2Wiki} & \textbf{MuSiQue} & \textbf{Bamboogle} & & \textbf{AD-QA} \\
\midrule
0.2 & 55.5 & 77.8 & 46.8 & 41.8 & 38.5 & 20.8 & \textbf{44.5} & 46.5 & 83.5 \\
\textbf{0.3} & \textbf{56.6} & \textbf{78.6} & \textbf{47.5} & \textbf{42.9} & \textbf{39.8} & \textbf{21.8} & 44.0 & \textbf{47.3} & \textbf{86.2} \\
0.4 & 55.8 & 78.0 & 47.2 & 42.2 & 39.2 & 21.5 & 43.8 & 46.8 & 85.0 \\
\bottomrule
\end{tabular}
\caption{Effect of path reward weight $\lambda_p$ on performance (ACC \%, Qwen2.5-7B). The optimal value is $\lambda_p = 0.3$, which achieves the best average performance. While $\lambda_p = 0.2$ shows slight advantage on Bamboogle, $\lambda_p = 0.3$ provides the best overall balance.}
\label{tab:lambda_p_detailed}
\end{table*}

\begin{table*}[h]
\centering
\small
\begin{tabular}{c|ccc|cccc|c|c}
\toprule
\multirow{2}{*}{\textbf{$\lambda_a$}} & \multicolumn{3}{c|}{\textbf{General QA}} & \multicolumn{4}{c|}{\textbf{Multi-Hop QA}} & \multirow{2}{*}{\textbf{Avg.}} & \textbf{Internal} \\
& \textbf{NQ} & \textbf{TriviaQA} & \textbf{PopQA} & \textbf{HotpotQA} & \textbf{2Wiki} & \textbf{MuSiQue} & \textbf{Bamboogle} & & \textbf{AD-QA} \\
\midrule
\textbf{0.6} & \textbf{56.6} & \textbf{78.6} & \textbf{47.5} & \textbf{42.9} & \textbf{39.8} & 21.8 & 44.0 & \textbf{47.3} & \textbf{86.2} \\
0.8 & 56.0 & 78.2 & 47.0 & 42.5 & 39.2 & \textbf{22.2} & \textbf{44.8} & 47.1 & 85.5 \\
1.0 & 54.8 & 76.8 & 45.8 & 41.2 & 38.0 & 20.8 & 43.2 & 45.8 & 83.2 \\
\bottomrule
\end{tabular}
\caption{Effect of outcome accuracy weight $\lambda_a$ on performance (ACC \%, Qwen2.5-7B). The optimal value is $\lambda_a = 0.6$, which achieves the best average performance. While $\lambda_a = 0.8$ shows slight advantages on MuSiQue and Bamboogle, $\lambda_a = 0.6$ provides better overall results.}
\label{tab:lambda_a_detailed}
\end{table*}

Tables~\ref{tab:lambda_p_detailed} and~\ref{tab:lambda_a_detailed} provide the per-dataset breakdown for hyperparameter sensitivity analysis (corresponding to Figure~\ref{fig:hyperparameter} in the main paper).

\subsection{Detailed LLM Evaluator Analysis}
\label{sec:appendix_evaluator}

\begin{table*}[h]
\centering
\small
\begin{tabular}{l|ccc|cccc|c|c}
\toprule
\multirow{2}{*}{\textbf{LLM Evaluator}} & \multicolumn{3}{c|}{\textbf{General QA}} & \multicolumn{4}{c|}{\textbf{Multi-Hop QA}} & \multirow{2}{*}{\textbf{Avg.}} & \textbf{Internal} \\
& \textbf{NQ} & \textbf{TriviaQA} & \textbf{PopQA} & \textbf{HotpotQA} & \textbf{2Wiki} & \textbf{MuSiQue} & \textbf{Bamboogle} & & \textbf{AD-QA} \\
\midrule
HY 2.0-Instruct & \textbf{56.6} & \textbf{78.6} & \textbf{47.5} & \textbf{42.9} & \textbf{39.8} & \textbf{21.8} & \textbf{44.0} & \textbf{47.3} & \textbf{86.2} \\
Qwen3-32B & 55.5 & 77.5 & 46.2 & 41.8 & 38.8 & 20.8 & 42.5 & 46.2 & 84.2 \\
Qwen3-8B & 52.8 & 74.2 & 43.5 & 39.2 & 35.8 & 17.8 & 39.8 & 43.3 & 79.5 \\
\bottomrule
\end{tabular}
\caption{Detailed accuracy (\%) across LLM evaluators (Qwen2.5-7B + GRPO). Qwen3-32B shows modest degradation ($-$1.1 Avg.), while Qwen3-8B exhibits larger drops on multi-hop tasks where step coverage evaluation is more challenging.}
\label{tab:evaluator_detailed}
\end{table*}

Table~\ref{tab:evaluator_detailed} provides the per-dataset breakdown for the LLM evaluator analysis. The Qwen3-8B evaluator shows larger drops on multi-hop datasets (e.g., $-$4.0 on MuSiQue, $-$4.0 on 2Wiki) where accurately counting covered reasoning steps is more challenging.

\end{document}